%% file: main.tex
\documentclass[runningheads]{llncs}

\usepackage[T1]{fontenc}

\usepackage{graphicx}

\usepackage[hidelinks]{hyperref} 
\usepackage{color}

\usepackage{amssymb}             
\usepackage{pifont, xcolor}      
\usepackage{cleveref}            
\RequirePackage{xspace}          
\usepackage{booktabs}            
\usepackage{orcidlink}           
\usepackage{capt-of}             

\makeatletter
\DeclareRobustCommand\onedot{\futurelet\@let@token\@onedot}
\def\@onedot{\ifx\@let@token.\else.\null\fi\xspace}
\def\eg{\emph{e.g}\onedot}

\def\etal{\emph{et al}\onedot}
\makeatother

\newcommand{\cmark}{\textcolor{green!60!black}{\ding{51}}} 
\newcommand{\xmark}{\textcolor{red!70!black}{\ding{55}}}   

\begin{document}

\title{OCCAM: Class-Agnostic, Training-Free, Prior-Free and
Multi-Class Object Counting}

\titlerunning{Training-Free, Prior-Free and
Multi-Class CAC}

\author{
Michail Spanakis\inst{1,2}\orcidlink{0009-0000-9160-6566} \and
Iason Oikonomidis\inst{2}\orcidlink{0000-0002-9503-3723} \and
Antonis Argyros\inst{1,2}\orcidlink{0000-0001-8230-3192} 
}

\authorrunning{M. Spanakis et al.}

\institute{
Computer Science Department, University of Crete, Heraklion, Greece
\and
Institute of Computer Science (ICS), 
Foundation for Research \& Technology – Hellas (FORTH), Heraklion, Greece\\
\email{\{mikespan,oikonom,argyros\}@ics.forth.gr}
}

\maketitle  

\begin{abstract} 
Class-Agnostic object Counting (CAC) involves counting instances of objects from arbitrary classes within an image. Due to its practical importance, CAC has received increasing attention in recent years. Most existing methods assume a single object class per image, rely on extensive training of large deep learning models and address the problem by incorporating additional information, such as visual exemplars or text prompts. In this paper, we present OCCAM, the first training-free approach to CAC that operates without the need of any supplementary information. Moreover, our approach addresses the multi-class variant of the problem, as it is capable of counting the object instances in each and every class among arbitrary object classes within an image. We leverage Segment Anything Model 2 (SAM2), a foundation model, and a custom threshold-based variant of the First Integer Neighbor Clustering Hierarchy (FINCH) algorithm to achieve competitive performance on widely used benchmark datasets, FSC-147 and CARPK. We propose a synthetic multi-class dataset and F1 score as a more suitable evaluation metric. The code for our method and the proposed synthetic dataset will be made publicly available at \url{https://mikespanak.github.io/OCCAM_counter}.

\keywords{Object Counting \and Class-Agnostic \and Multi-Class \and Training-Free \and Exemplar-Free \and Segment Anything Model (SAM) \and FINCH.}
\end{abstract}

\input{sections/1_intro}

\input{sections/2_related_work}

\input{sections/3_methodology}

\input{sections/4_experiments}

\input{sections/5_conclusions}




\bibliographystyle{splncs04}
\bibliography{mybibliography}

\end{document}

%% file: sections/1_intro.tex
\section{Introduction}
\label{chapter:introduction}

Object counting is a natural and intuitive ability. Counting fundamentals emerge in humans from a very young age - often even before the development of speech~\cite{wynn1992addition}.
But what happens when the number of objects becomes large? Counting becomes tedious and repetitive. Despite its apparent simplicity, counting remains a cognitively demanding task. It follows object recognition, localization, and differentiation, making it especially challenging for computers.

\begin{figure}
  \centering
  \includegraphics[width=0.60\linewidth]{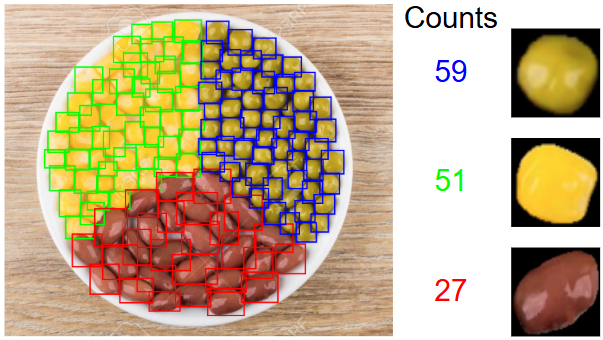}
  \caption{We propose a class-agnostic multi-class object counting approach that takes as input an RGB image and outputs a count of object instances per detected class.}
  \label{fig:fig1}
\end{figure}

Despite these challenges, numerous approaches have been developed for counting objects in images. Traditional computer vision techniques were used in the initial methods, \eg ~\cite{lempitsky2010learning}. In recent years, class-specific object counting methods have achieved significant progress utilizing prior knowledge regarding the target object class due to advances in neural networks and deep learning. Specialized models have been developed for counting humans in crowded scenes~\cite{deng2024deep}, vehicles~\cite{hsieh2017drone}, animals~\cite{arteta2016counting}, and other categories of objects in various domains~\cite{rahnemoonfar2017deep,zavrtanik2020segmentation,zheng2024rethinking}. 
These counting models can reliably count a large number of objects belonging to a single predefined class. As long as sufficiently large annotated training datasets for each object class are available, there is no inherent restriction on the target object class. Collecting and annotating these datasets is often prohibitively expensive. Furthermore, most existing approaches require training a separate model for each category of objects, which limits their scalability and flexibility. These limitations have led to the emergence of a more generalized and demanding research direction: class-agnostic object counting~\cite{lu2019class}.  

To address the aforementioned limitations, approaches have been developed to count objects from arbitrary object classes. These are typically trained on  images including several objects in various object categories to learn the underlying notion of similarity between object instances. However, most existing methods assume the presence of only a single object class per image. As a result, they rely on additional input, such as visual exemplars or text prompts, to specify which objects should be counted. Some studies have introduced interactive counting frameworks~\cite{arteta2014interactive,huang2023interactive} that incorporate human feedback into the loop to refine the predictions. Other approaches~\cite{ranjan2022exemplar} count only the object class with the highest number of instances. Although these methods represent important progress in the field, they still face challenges in terms of generalization and practical applicability to diverse real-world scenarios.

Recent approaches omit the dependence on extensive training by leveraging pre-trained, general-purpose vision models, such as SAM~\cite{kirillov2023segment} and DINOv2~\cite{oquab2023dinov2}. In addition, some methods~\cite{ranjan2022exemplar,hobley2022learning} operate without requiring any priors. A few methods revoke the assumption of a single object class per image~\cite{pelhan2024dave,jeon2025mutually,xu2025learning} or aim to count instances from multiple classes~\cite{hobley2024abc,mondal2025omnicount}. Our method, Object Counting: Class-agnostic, Assumption-minimal, Multi-class (OCCAM), advances this research direction by imposing, to the best of our knowledge, the fewest assumptions to date. Specifically, our method is: (1)~\textbf{Training-free}: it does not depend on some kind of training procedure, allowing immediate deployment out of the box; (2)~\textbf{Prior-free}: only requires a plain RGB image, since additional information implicitly defines the target object category a priori; (3)~\textbf{Multi-class}: it is capable of counting multiple object categories within an image in a single pass. To achieve the above, OCCAM leverages SAM2~\cite{ravi2024sam} and FINCH~\cite{sarfraz2019efficient}. 

Extensive experimental evaluation demonstrates the effectiveness and generalizability of our approach in several datasets. To ease the evaluation of the mullticlass aspects of OCCAM, we introduce a new synthetic dataset. Moreover, we discuss the evaluation metrics adopted in the relevant literature up to now and we argue that the F1-score is more suitable for evaluating CAC.  OCCAM achieves competitive and, in some cases, state-of-the-art performance across standard metrics. There is currently no other CAC method being both Prior-free and Training-free. All these properties combined, make OCCAM truly class-agnostic, ideal for fully automated object counting systems and position it as the most general solution available.

In summary, the key contributions of our work are:
\vspace{-0.20cm}
\begin{itemize}
    \item We introduce the first Class-Agnostic object Counting (CAC) method that is entirely training-free, prior-free, and capable of counting object instances in multiple classes within an image.    
    \item We introduce a synthetic multi-class dataset for object counting, constructed by stitching together an arbitrary number of real-world single-class images. 
    \item We propose F1 score as a more suitable and unbiased evaluation metric for CAC. This enhances the validity of the quantitative evaluation.
\end{itemize}

%% file: sections/2_related_work.tex
\section{Related Work}
\label{chapter:related_work}

Class-Agnostic object Counting (CAC) generalizes object counting across arbitrary categories, reducing reliance on class-specific training. Existing CAC methods can be grouped by (i) whether they depend on priors (\eg exemplars, points or text prompts), (ii) whether they require dedicated training, and (iii) whether they assume a single dominant class per image or support multi-class counting.

\noindent{\bf  Prior-Based CAC.}
Most CAC methods use priors to specify what to count during training and inference. Exemplar-based methods leverage the strongest guidance via 1--3 cropped instances from the given image. 
Counting is typically cast as matching or similarity learning, from the first self-similarity formulation in~\cite{lu2019class} proposing GMN, to few-shot regression-based FamNet~\cite{ranjan2021learning}, and BMNet/BMNet+~\cite{shi2022represent} which jointly learn feature representation and a similarity metric. Nguyen \etal ~\cite{nguyen2022few} proposed Counting-DETR, an uncertainty-aware detector trained via pseudo labels, while SAFECount~\cite{you2023few} refines separation in crowded scenes through similarity-aware comparison and feature fusion. Text-prompted CAC improves generality by weakening the prior, enabling counting from class labels~\cite{xu2023zero}. Recently, multi-modal methods have been developed. CounTR~\cite{liu2022countr} is a transformer-based model that leverages attention to compare image patches with exemplars and LOCA~\cite{djukic2023low} uses an iterative Prototype Extraction module to fuse exemplar shape and appearance with image features. PseCo~\cite{huang2024point} combines SAM~\cite{kirillov2023segment} and CLIP~\cite{radford2021learning} in a point--segment--count pipeline, while MAFEA~\cite{jeon2025mutually} mitigates target confusion via cross-attention. In DAVE~\cite{pelhan2024dave} density-based candidate generation is combined with a detection-based verification stage to improve localization and outlier removal. A single-stage counter GeCo~\cite{pelhan2024novel} unifies detection, segmentation, and counting. COUNTGD~\cite{amini2024countgd} utilizes GroundingDINO~\cite{liu2024grounding} fusing text and visual prompts via attention.  

\noindent{\bf  Prior-Free CAC.}
A smaller body of work~\cite{ranjan2022exemplar,hobley2022learning,wu2024gca,zgaren2025save} avoids external priors and instead exploits self-similarity and internal feature relationships. RepRPN-Counter~\cite{ranjan2022exemplar}, the first exemplar-free CAC method, discovers repeating objects via an RPN and estimates counts through density prediction. RCC~\cite{hobley2022learning} frames counting as repetition recognition under weak supervision. GCA-SUN~\cite{wu2024gca} directly predicts density maps using gated context-aware modulation, feature selection, and multi-scale fusion, while SAVE~\cite{zgaren2025save} uses detection-based features, self-attention over visual embeddings, and count regression. While prior-free approaches are suitable for automation, the ambiguity of what to count becomes more prominent, making the single-class-per-image assumption essential. Furthermore, some multimodal methods~\cite{liu2022countr,djukic2023low,pelhan2024dave,pelhan2024novel,jeon2025mutually} can operate without additional information, typically trading off accurate object counting for generality.

\noindent{\bf Training-Free CAC.}
Most CAC methods still require extensive task-specific training of advanced deep learning models and large-scale datasets making them resource-intensive. Training-free methods~\cite{ma2023can,shi2024training,ting2024tfcounter,lin2025simple,pacini2025countingdino,huang2025training} aim to overcome these limitations by leveraging foundation models. Ma \etal~\cite{ma2023can} were the first to use SAM~\cite{kirillov2023segment} exploiting exemplar-guided mask generation and cosine-similarity-based feature averaging for counting. TFOC~\cite{shi2024training} also uses SAM, formulating counting as prior-guided segmentation integrating similarity, segment, and semantic priors.   TFCounter~\cite{ting2024tfcounter} utilizes a dual-point prompt strategy and a context-aware similarity module, while in~\cite{lin2025simple} instance-level counting with improved point prompting and semantically enriched features is performed by SAM. CountingDINO~\cite{pacini2025countingdino} relies on fully unsupervised DINOv2~\cite{oquab2023dinov2} features to extract latent object prototypes and generate similarity-based density maps. Count validity is improved by combining SAM-based spatial proposals with EVA-CLIP–based~\cite{sun2023eva} semantic filtering in~\cite{huang2025training} ensuring that only spatially and semantically relevant object instances are counted.

\noindent{\bf  Multi-Class CAC.}
Real-world images typically contain multiple object categories. A few methods~\cite{pelhan2024dave,jeon2025mutually,xu2025learning} have been developed to utilize priors accordingly to count object instances of the target class despite the presence of multiple categories. Xu \etal~\cite{xu2025learning} combine exemplar-based segmentation with pseudo-labeled masks to mitigate multi-class over-counting and highlight the trade-off between handling multiple classes and single-class counting accuracy.

Only recent works are capable of counting each object instance belonging in each and every class within an image. ABC123~\cite{hobley2024abc}, the first multi-class CAC method, is a transformer-based density regression approach trained on the synthetic MCAC dataset to simultaneously count multiple object types without priors. Each object class can have up to 300 object instances and at most 4 different categories can be counted per image. OCCAM overcomes these limitations despite being prior-free; it can count objects belonging to an arbitrary number of different classes without an upper bound on their instances. OmniCount~\cite{mondal2025omnicount} is training-free and detection-based, using SAM for instance segmentation relying on semantic and geometric priors to count multiple categories in a single pass. However, OCCAM does not rely on provided priors. Overall, OCCAM is the most general CAC method to date and there is no direct competitor. Every comparison should be made taking into consideration the above differences.

%% file: sections/3_methodology.tex
\section{Methodology}
\label{chapter:methodology}

\begin{figure}[t]
    \centering
    \includegraphics[width=1.0\linewidth,trim={0 5.0cm 0 5.0cm},clip]{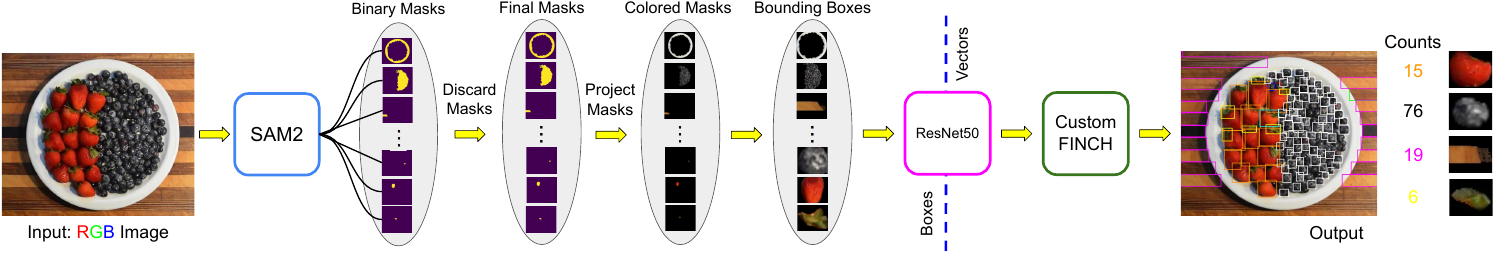}
    \caption{Given an RGB image as input, a dense grid of seed points is constructed, which are passed to SAM2 to produce a collection of binary masks. These masks are then filtered to obtain a refined set of candidate object masks. Each candidate object mask is used to extract an aspect-ratio-preserving bounding box region, which is passed through ResNet50 to produce a feature vector. These vectors are then grouped using a custom threshold-based variant of FINCH algorithm, resulting in a final set of clusters.}
    \label{fig:Pipeline}
\end{figure}

We aim to address training-free, multi-class, class-agnostic object counting without using priors. In order to achieve this, our method, Object Counting: Class-agnostic, Assumption-minimal, Multi-class (OCCAM), is based on two key components: the Segment Anything Model 2 (SAM2)~\cite{ravi2024sam} and a threshold-based variant of the First Integer Neighbor Clustering Hierarchy (FINCH) algorithm~\cite{sarfraz2019efficient} that we propose. Figure~\ref{fig:Pipeline} illustrates the proposed approach. 

\noindent{\bf Object Mask Generation.}
\label{sec:maskgen}
We use the Segment Anything Model 2 (SAM2)~\cite{ravi2024sam} to obtain masks per object instance in the image without prior knowledge of object categories, number, or size. Specifically, since the RGB image $I \in \mathbb{R}^{H \times W \times 3}$ is our sole form of input, we densely sample a custom canonical grid of possible seed points, $p_i$, over $I$ and use them as point-prompts for SAM2. Each $p_i$ may correspond to a distinct object of interest. The spacing between the seed points is chosen empirically and remains constant across all images, resulting in a regular grid whose size depends on the dimensions of $I$. For each $p_i$, SAM2 produces up to three binary masks $M_i$ without requiring any form of additional information.

\noindent{\bf Object Mask Postprocessing.}
\label{sec:maskproc}
Further processing of the raw binary masks $M_i$ is necessary, since there is no guarantee that each mask produced by SAM2 corresponds to a distinct and unique object instance. There are two main issues: (1) different seed points may produce the same, or a slightly different, binary mask, and (2) one to three masks could be produced by SAM2 from the same seed point. Therefore, each pixel may not be uniquely assigned to a single mask, which can adversely affect the final object count.

In order to resolve these issues, we proceed to mask filtering. Regarding the first issue, duplicate masks are discarded. Specifically, masks are considered ``duplicates'' if their Intersection over Union (IoU) value exceeds an empirically determined threshold, in which case only one mask is retained arbitrarily. Enforcing SAM2 to produce a single mask per seed point would not resolve the second issue, as the target class is not specified a priori due to the lack of additional information. We rely on our proposed FINCH variant to overcome this problem. However, we try to mitigate it by applying more filtering rules. If a mask consists of multiple connected components, only the major one is kept, unless it constitutes a duplicate with a different mask produced independently. This is important because the produced masks often contain a couple of distinct objects, each of which could be a separate target object. As a result, only a single object instance is contained in each mask, any occlusive object is excluded and the size of the segmented object can be estimated more accurately.

The estimated size of each object instance is utilized in the following filtering rules. Final binary masks that are too big, relatively to the image size, are discarded, since they may correspond to the background or be a super-set of other masks. Additionally, masks with a single-pixel width or height are discarded, as they are likely artifacts. Overall, these filtering rules eliminate some binary masks, $M_i$, resulting in the final set of candidate object masks, $\hat{M}_j$.

\noindent{\bf Multiscale Refinement Strategy.}
\label{sec:scalingpar}
In case the final number of candidate object masks found within the image is too small, a simple scaling scheme, similar to a test-time normalization technique used in \cite{liu2022countr}, is deployed. The given image is split into a non-overlapping canonical $3\times3$ grid and the entire methodology is recursively applied to each of the produced sub-images, independently. Filtered binary masks produced from each of the nine sub-images are accumulated to form the final set of candidate object masks $\hat{M}_j$. This strategy enables the model to detect smaller objects that may be overlooked in the original image. If no additional objects are detected using this scaling scheme, we conclude that this image contains very few objects. Through this mechanism, our method can differentiate between uniform images, or empty scenes, and images containing numerous tiny objects.

\noindent{\bf Feature Extraction.}
\label{sec:featextr}
The next step is to cluster the final set, $\hat{M}_j$, of candidate object masks. First, each retained mask, $\hat{M}_j$, is applied to the original image to extract the corresponding colored region, and tight bounding boxes are formed. In order to avoid any size-related bias, each bounding box is resized to a predetermined size, while preserving the original aspect ratio of the depicted object. Any residual space of each bounding box is zero-padded. Consequently, each of these aspect-ratio-preserving bounding boxes is color-normalized and passed through a deep model for a standardized representation to be extracted. To achieve this, the standard ResNet-50~\cite{he2016deep} pre-trained on ImageNet~\cite{deng2009imagenet} excluding its last fully connected layer is utilized. Therefore, this procedure results in a 2048-dimensional feature vector $f_j$, $\mathbf{f}_j \in \mathbb{R}^{2048}$, produced for each bounding box.

\noindent{\bf Threshold-based FINCH Variant.}
\label{sec:thresFINCH}
A custom threshold-based variant of the First Integer Neighbor Clustering Hierarchy (FINCH) algorithm~\cite{sarfraz2019efficient} is proposed to properly group the extracted feature vectors, $f_j$, into an arbitrary number of clusters. Each of these clusters should ideally correspond to a distinct object category in the image.

The key idea of FINCH is to iteratively group vectors which are the closest neighbors to each other, based on a chosen distance metric. The closest-neighbor relation though can be asymmetrical; a vector $v_i$ may be the closest neighbor of another vector $v_j$, but this does not necessarily imply that $v_j$ is also the closest neighbor of $v_i$. This may occur as $v_i$ may have a closer neighbor in a different direction. FINCH bypasses this issue by enforcing symmetry, effectively allowing multiple closest neighbors. Hence, in our example, $v_i$ is also marked as the closest neighbor of $v_j$, resulting in a situation where $v_j$ has two ``closest neighbors''. Furthermore, the absolute value of the distance is not taken into consideration, allowing the algorithm to remain completely parameter-free. Finally, the minimum possible size of a cluster is two, which could cause various issues in our case, such as the adverse grouping of the feature vector corresponding to the background of the image.

In our approach, the following modifications are applied to the FINCH algorithm. An  empirical threshold is introduced in each step of the FINCH algorithm. Thus, two vectors will not be grouped together if the distance between them is greater than this threshold, despite their partial closest-neighbor relation. In all of our experiments, the Euclidean distance is used in the 2048-dimensional feature space. For the first iterations, the corresponding threshold decreases in each step, while it remains constant for the later iterations, balancing intra-class variance and inter-class discrimination. This is a simple but effective way to prevent the enforcement of symmetry and resolve the aforementioned issues. 

Initially, each vector forms a cluster by itself. After the first iteration, the centroids of the formed clusters are estimated, and the distances among clusters are computed based on these centroids. The clusters which are, at least, partial closest neighbors and their centroids are at a distance less than the defined threshold are merged. The iterative procedure is terminated when there are no more clusters left to be merged. Compared to the original FINCH method, this constitutes a different termination condition, as FINCH stops producing hierarchical clustering partitions when the next partition would include every given element. Through this approach, we eliminate the need to devise a non-trivial automatic procedure for selecting the optimal clustering level from the hierarchical partitions generated by the default FINCH algorithm. 

The output of our proposed threshold-based FINCH variant is the resulting clusters, $C_k$, each of which corresponds to a number of candidate object masks present in the input image. As far as the evaluation procedure is concerned, each $C_k$ is matched to the ground-truth one according to which it shares the largest number of object instances with.

%% file: sections/4_experiments.tex

\begin{table}[t]
    \centering
    \caption{Multi-Class Methods on FSC-147 Test Set (best results in \textbf{bold}). Results reported below the double line include images with $\leq 300$ objects.}
        \begin{tabular}{lccrcc}
            \toprule
            Model     & Info      & MAE $\downarrow$ & RMSE $\downarrow$ & NAE $\downarrow$ & SRE $\downarrow$ \\
            \midrule
            Mean      & - & 47.55 & 147.67 & - & - \\	
            Median    & - & 47.73 & 152.46 & - & - \\
            \midrule
            OmniCount \cite{mondal2025omnicount}       & Exemplars & 18.63 & 112.98 & \textbf{0.14} & 2.99 \\
            OmniCount  \cite{mondal2025omnicount}      & Points    & 19.24 & 115.27 & 0.25 & 3.21 \\
            OmniCount \cite{mondal2025omnicount}       & Text      & 21.46 & 133.28 & 0.32 & \textbf{0.39} \\
            OCCAM-S	  & -         & \textbf{16.92} & \textbf{110.83} & 0.22 & 3.38 \\
            \midrule
            \midrule
            ABC123 \cite{hobley2024abc}   & -         & 11.75 & 33.41  & \textbf{0.11} & \textbf{2.38} \\
            OCCAM-S	($\leq$ 300)  & -         & \textbf{11.29} & \textbf{25.17} & 0.21 & 2.68 \\
            \bottomrule
        \end{tabular}
    \label{tab:1}
\end{table}


\section{Experiments}
\label{chapter:experiments}

\noindent{\bf Datasets.}
{\sl\underline{FSC-147 Dataset}:}
The FSC-147 dataset~\cite{ranjan2021learning} is the benchmark for few-shot, class-agnostic object counting, consisting of 6135 human-annotated images from 147 object classes with non-overlapping object categories in train/val/test splits. The test set contains 1190 images from 29 diverse classes, from sea shells to stamps, with per-instance center annotations and three exemplar bounding boxes per image. The average count is 56 objects per image. Since our method is training-free and prior-free, only the test split is used, without relying on exemplars or annotations. A key limitation of FSC-147 is the imbalanced class distribution, and the single-class-per-image assumption, which is not realistic and does not hold for a great number of images. The common heuristic that the most frequent class is the annotated one is not always valid. Thus, FSC-147 is less suitable for evaluating our method.\\
{\sl \underline{Synthetic Multi-class Test Set:}} We construct a synthetic multi-class test set by stitching together randomly selected images from the FSC-147 test split. Three dataset variants are generated to assess robustness, each containing 100 images composed of 1–10 sub-images containing a different and distinct object class.
Images are stitched side by side using normalized height, forming grids of at most two columns, without resizing; unused regions are zero-padded. The class composition and instance counts are unconstrained, resulting in images with high object densities. All of the above make the proposed dataset challenging and suitable for multi-class counting.\\
{\sl \underline{Real Multi-class Test Set:}} A recently proposed dataset~\cite{xu2025learning} has been acquired by privately contacting the respected authors. 
It contains 453 real-world images, each depicting 1–3 object classes from 76 different object categories with possible inter-class overlap. For consistency with FSC-147, three exemplars per class and center annotations per instance are provided. Its major advantage is that object instances are annotated across all classes present in each image, making this dataset well suited for evaluating multi-class CAC methods.\\

\noindent{\sl \underline{CARPK Dataset:}} CARPK~\cite{hsieh2017drone} is a counting benchmark consisting of 1448 aerial drone images, with cars as the only annotated class using tight bounding boxes; 459 images are used for testing. In the CAC field, it is utilized to examine the ability of the proposed approaches to generalize adequately to domain-specific counting tasks.


\begin{table*}[t]
\centering
\begin{minipage}[t]{0.515\textwidth}
    \centering
    \small
    \captionof{table}{Training-Free Methods on FSC-147 Test Set (best results in \textbf{bold}).}
        \begin{tabular}{llcr}
        \toprule
        Model        & Info      & MAE $\downarrow$ & RMSE $\downarrow$ \\
        \midrule
        CAN SAM? \cite{ma2023can}    & 3-shot & 27.97 & 131.24 \\
        CountingDINO \cite{pacini2025countingdino} & 3-shot & 20.93 & 71.37  \\
        TFOC \cite{shi2024training}        & 3-shot & 19.95 & 132.16 \\
        ValidCounter \cite{huang2025training} & 3-shot & 19.33 & 133.33 \\
        TFCounter \cite{ting2024tfcounter}   & 3-shot & 18.56 & 130.59 \\
        A-Simple-But \cite{lin2025simple} & 3-shot & \textbf{12.26} & \textbf{56.33}  \\
        \midrule
        CountingDINO \cite{pacini2025countingdino} & 1-shot    & 37.73 & 131.71 \\
        CAN SAM? \cite{ma2023can}    & 1-shot    & 33.53 & 142.28 \\
        A-Simple-But \cite{lin2025simple} & 1-shot    & \textbf{15.15} & \textbf{69.16}  \\
        \midrule
        TFOC \cite{shi2024training}        & Text      & 24.79 & 137.15 \\
        A-Simple-But \cite{lin2025simple} & Text      & \textbf{23.59} & \textbf{113.60} \\
        \midrule
        \textbf{OCCAM-S}	     & \textbf{-}         & \textbf{16.92} & \textbf{110.83}\\
        \bottomrule
        \end{tabular}
    \label{tab:2}
\end{minipage}
\hfill
\begin{minipage}[t]{0.40\textwidth}  
    \centering
    \small
    \captionof{table}{Prior-Free Methods on FSC-147 Test Set (best results in \textbf{bold}). Scores for RepRPN-Counter \cite{ranjan2022exemplar} are for ``Top 5''.}
        \begin{tabular}{lrr}
        \toprule
        Model          & MAE $\downarrow$ & RMSE $\downarrow$ \\
        \midrule
        RepRPN-C \cite{ranjan2022exemplar} & 26.66 & 129.11 \\
        RCC \cite{hobley2022learning}           & 17.12 & 104.53 \\
        LOCA \cite{djukic2023low}          & 16.22 & 103.96 \\
        DAVE \cite{pelhan2024dave}          & 15.14 & 103.49 \\
        CounTR \cite{liu2022countr}        & 14.71 & 106.87 \\
        GCA-SUN \cite{wu2024gca}       & 14.00 & 92.19  \\
        GeCo \cite{pelhan2024novel}          & 13.30 & 108.72 \\
        MAFEA \cite{jeon2025mutually}         & 13.23 & 105.99 \\
        SAVE \cite{zgaren2025save}       & \textbf{8.92}  & \textbf{80.39} \\
        \midrule
        OCCAM-S           & 16.92 & 110.83 \\
        \bottomrule
        \end{tabular}
    \label{tab:3}
\end{minipage}
\end{table*}


\noindent{\bf Implementation Details.}
According to our experiments, two slightly different configurations of our methodology give the best results depending on whether the given dataset is multi-class or not. Both configurations rely on ``SAM 2.1 Large''~\cite{ravi2024sam}. A seed point is set per $10$ pixels and an IoU threshold of $0.1$ is used. There are two differences: (1)~the size of the constructed bounding boxes, resized to $224\times224$ and $500\times500$ for single-class and multi-class, respectively, and (2)~the absolute value of the thresholds used in our custom FINCH variant. There is a decreasing threshold scheme in each of the first three iterations, $12.0$, $9.0$, $7.75$ in single-class cases and $5.0$, $4.0$, $3.0$ in multi-class.
Therefore, we end up with two configurations, OCCAM-S and OCCAM-M, for single-class and multi-class scenarios, respectively.

\noindent{\bf Evaluation Metrics.}
Our method is evaluated using the most widely adopted metrics in the CAC literature: Mean Absolute Error (MAE)
and Root Mean Squared Error (RMSE)
as originally introduced in~\cite{lu2019class,ranjan2021learning}. 
However, the significance of an absolute counting error depends heavily on the number and appearance of the present ground-truth objects. 
To account for this, Normalized Absolute Error (NAE) 
and Squared Relative Error (SRE),
which normalize the error relative to the number of objects in each image, are also reported. Since our approach is multi-class, the adjusted versions, introduced in \cite{hobley2024abc}, are employed.
These metrics can be misleading: a perfect numerical score does not ensure correct instance matching, as False Positives and False Negatives may cancel out. To mitigate this, we adopt Precision, Recall and the F1 Score which is unbiased. These are well-established in related domains and provide a more reliable and interpretable assessment of counting performance. They are also used in the foundational paper~\cite{lu2019class} and their relevance is further emphasized in~\cite{huang2025training}, a very recent work concurrent with ours.


\begin{table}[t]
\centering
\caption{Precision, Recall and F-1 Score on FSC-147 Test Set (best results in \textbf{bold}). All other methods use exemplars. Below the double line, the approaches are Training-free. Only our approach is Prior-Free. NAE, SRE, Precision, Recall and F-1 scores for other methods are taken from ValidCounter~\cite{huang2025training}.}
\resizebox{\columnwidth}{!}{
    \begin{tabular}{lcrcrccc}
        \toprule
        Model        & MAE $\downarrow$  & RMSE $\downarrow$  & NAE $\downarrow$  & SRE $\downarrow$ & Precision $\uparrow$ & Recall $\uparrow$ & F-1 $\uparrow$ \\
        \midrule
        FamNet+ \cite{ranjan2021learning}      & 22.08 & 99.54  & 0.44 & 6.45 & 0.66      & 0.67   & 0.57 \\
        BMNet+ \cite{shi2022represent}      & 14.62 & 91.83  & 0.27 & \textbf{6.19} & 0.66      & 0.68   & 0.67 \\
        CounTR \cite{liu2022countr}      & 11.95 & 91.23  & \textbf{0.23} & 7.44 & \textbf{0.79}      & \textbf{0.78}   & \textbf{0.78} \\
        LOCA \cite{djukic2023low}        & \textbf{10.79} & \textbf{56.97}  & 0.26 & 7.96 & 0.78      & \textbf{0.78}   & \textbf{0.78} \\
        \midrule
        \midrule
        TFOC \cite{shi2024training}        & 19.95 & 132.16 & 0.29 & 3.80 & 0.79      & \textbf{0.82}   & 0.80 \\
        ValidCounter \cite{huang2025training} & 19.33 & 133.33 & \textbf{0.20} & 3.43 & 0.87      & 0.77   & 0.82 \\
        \midrule
        OCCAM-S      & \textbf{16.92} & \textbf{110.83} & 0.22 & \textbf{3.38} & \textbf{0.89}      & \textbf{0.82}   & \textbf{0.83} \\
        \bottomrule
    \end{tabular}
}
\label{tab:4}
\end{table}



\begin{figure*}[t]
\centering
\newcommand{\imgH}{2.8cm}

\begin{tabular}{lrc}
\includegraphics[height=\imgH]{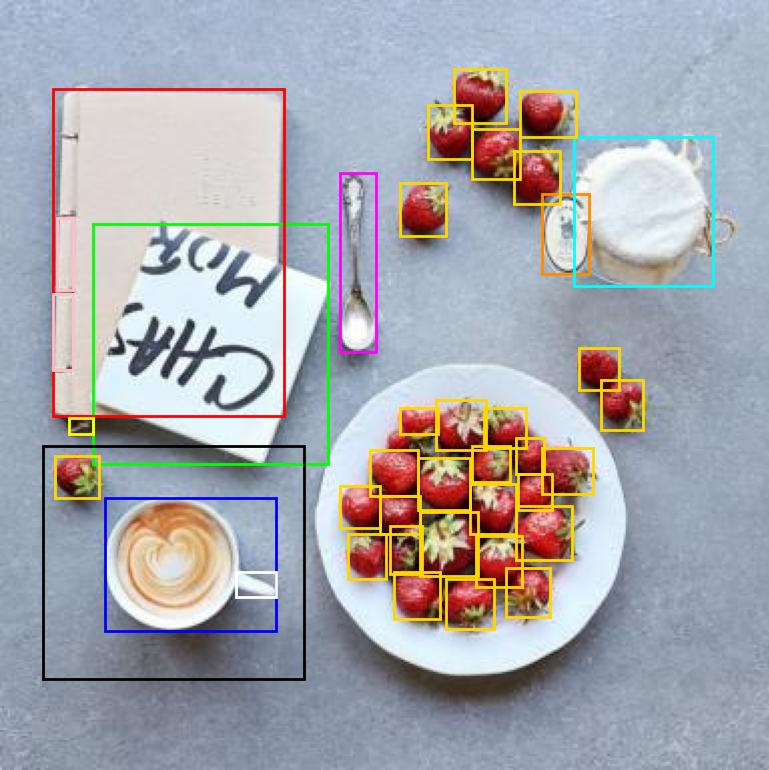} &
\includegraphics[height=\imgH]{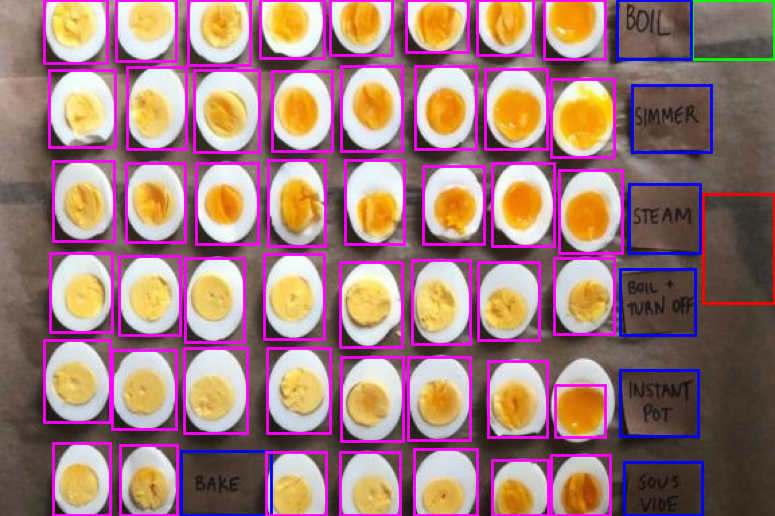} &
\includegraphics[height=\imgH]{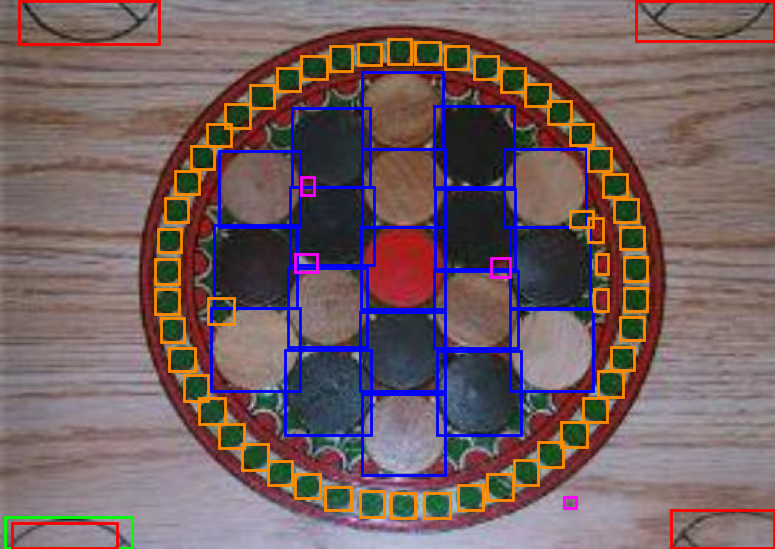} \\[2mm]

\includegraphics[height=\imgH]{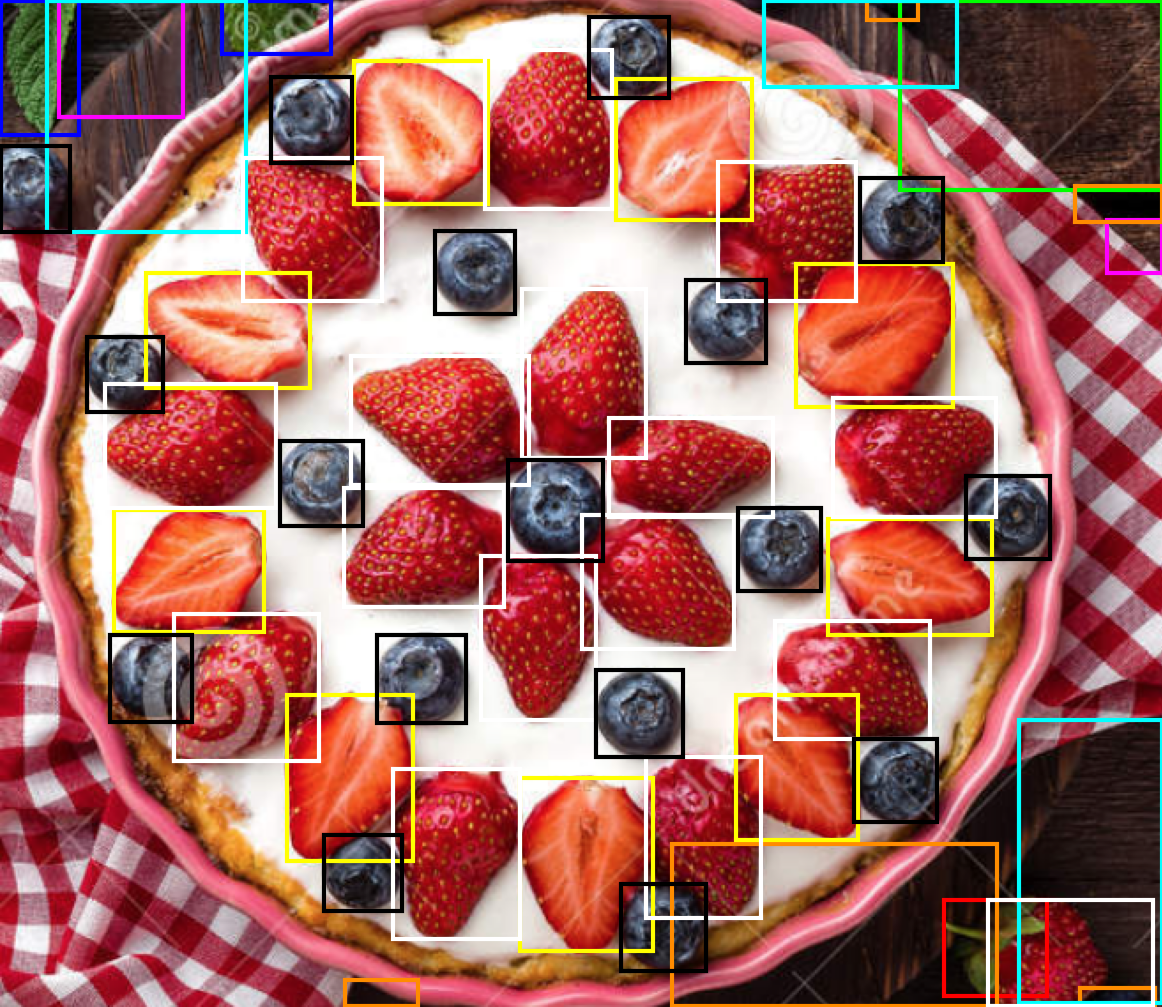} &
\includegraphics[height=\imgH]{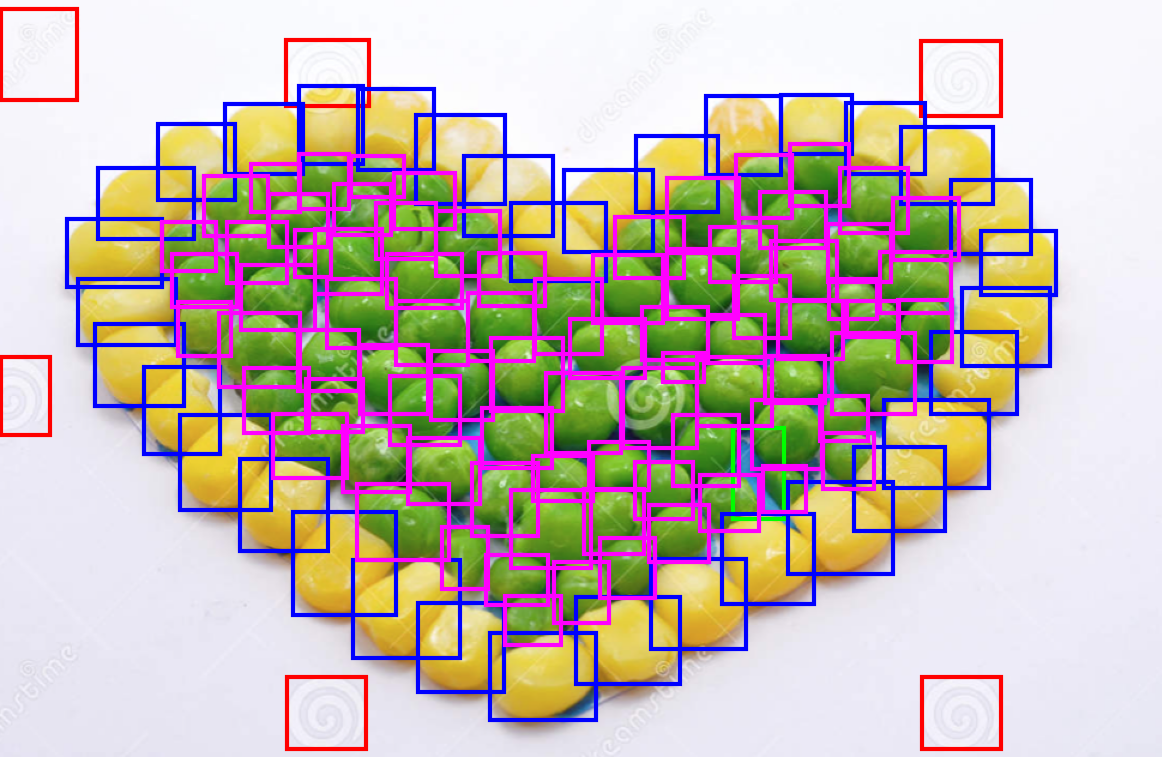} &
\includegraphics[height=\imgH]{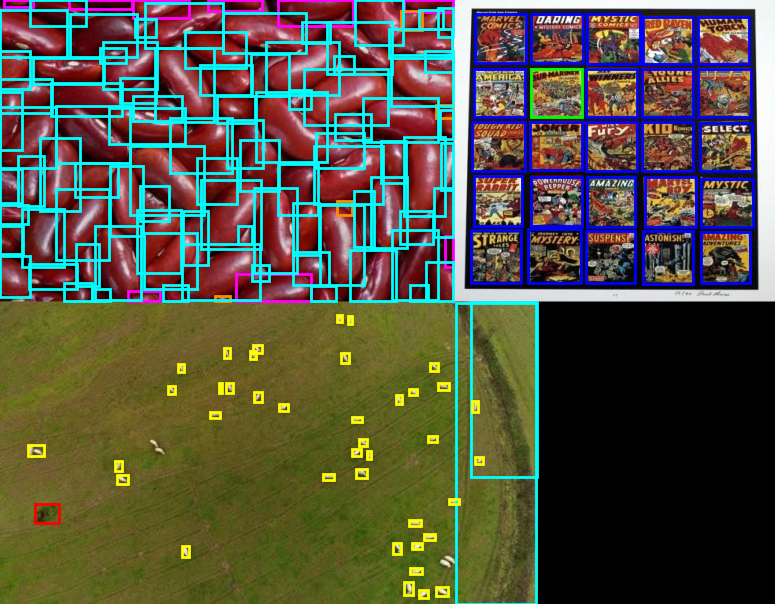} \\[2mm]

\includegraphics[height=\imgH]{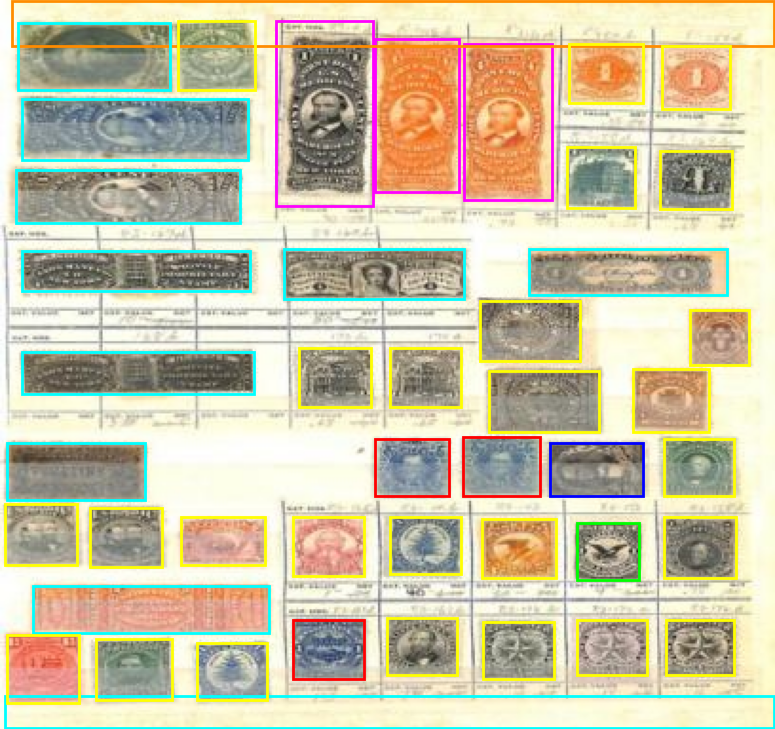} &
\includegraphics[height=\imgH]{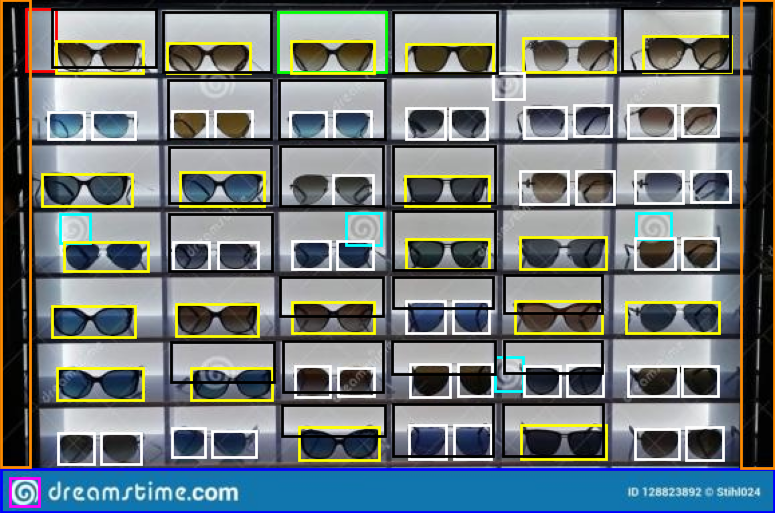} &
\includegraphics[height=\imgH]{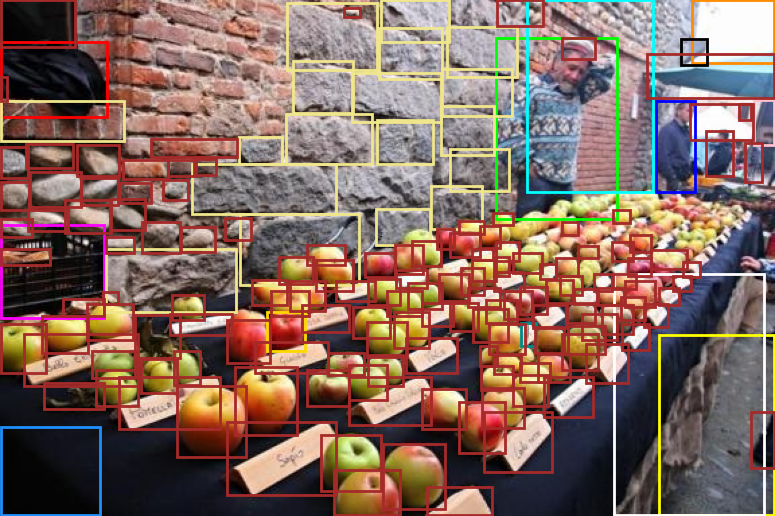}
\end{tabular}

\caption{Qualitative Results from the FSC-147 Test Set (first row) and the Real Multi-Class Test Set (second row). Failure Cases shown in the third row.}
\label{fig:qualres}
\end{figure*}


\noindent{\bf Quantitative Results.}
Despite not having a direct competitor due to its parsimonious assumptions, OCCAM is competitive with other multi-class and training-free approaches, while lacking behind from other prior-free approaches, each of which relies on training. Indicative of its success are the high Precision, Recall and F-1 scores. It generalizes well on the CARPK dataset and sets a new state-of-the-art MAE for the Real Multi-class Test Set. According to the obtained results, OCCAM performs remarkably well and even exhibits state of the art performance in certain cases.

In \cref{tab:1}, our approach is compared with the mean and median baselines and the few other multi-class methods. Our approach, outperforms the baselines by a significant margin and OCCAM-S achieves better MAE and RMSE scores compared to OmniCount~\cite{mondal2025omnicount}, our main competitor, without utilizing any priors. OCCAM-S achieves better MAE and RMSE scores than ABC123 and has competitive NAE and SRE despite the fact that it is training-free. In addition, the impact of a few images of the FSC-147 Test Set which contain more than 300 objects becomes evident by their exclusion.

In \cref{tab:2}, our approach is compared with other training-free methods using the FSC-147 Test Set. Each of the other methods is single-class and utilizes some sort of prior. However, we surpass most methods in MAE.

In \cref{tab:3}, our approach is compared with other prior-free methods using the FSC-147 Test Set. Given that our approach is the only multi-class and training-free method, our results are relatively good.

In \cref{tab:4}, our approach is compared with other methods regarding Precision, Recall and F1-Score using the FSC-147 Test Set. All other methods are single-class and utilize exemplars as prior information. Despite making the least assumptions, our approach achieves competitive scores, having state-of-the-art Precision, Recall and F1-Score.

In \cref{tab:5}, our approach is compared with other CAC methods in the Real Multi-Class Test Set~\cite{xu2025learning}. OCCAM-S achieves competitive MAE and RMSE with the other approaches trained solely using the FSC-147 training set. OCCAM-M even achieves state-of-the-art MAE. Precision, Recall, and F1-score of OCCAM-M are $0.90$, $0.81$, and $0.83$, respectively (not shown in the table).

In \cref{tab:6}, our approach is compared with other CAC methods using the CARPK dataset. We managed to achieve competitive scores across all metrics. Our OCCAM-S configuration has state-of-the-art Precision, $0.90$, Recall, $0.89$, and F-1 Score, $0.89$, the second best method being the ValidCounter \cite{huang2025training} scoring $0.87$, $0.84$, and $0.86$, respectively (not shown in the table).

In \cref{tab:7}, our approach is evaluated on our proposed Synthetic Multi-class Test Set.
Our OCCAM-M configuration achieves better results overall, as expected for a multi-class dataset.


\begin{table*}[t]
\centering
\begin{minipage}[t]{0.40\textwidth}
    \centering
    \small
    \captionof{table}{Methods on Real Multi-Class Test Set (best results in \textbf{bold}). Above the double line, methods use only the FSC-147 training set, while below also use a synthetic dataset~\cite{xu2025learning}.}
        \begin{tabular}{lrccc}
        \toprule
        Model & MAE $\downarrow$  & RMSE $\downarrow$ \\
        \midrule
        FamNet \cite{ranjan2021learning}                   & \textbf{13.03} & \textbf{20.28} \\
        FamNet+ \cite{ranjan2021learning}                  & 19.42 & 39.78 \\
        BMNet+ \cite{shi2022represent}                   & 25.55 & 40.35 \\
        SAFECount \cite{you2023few}                & 23.57 & 40.99 \\
        \midrule
        OCCAM-S                  & 16.82 & 42.21 \\
        \midrule
        \midrule
        FamNet \cite{ranjan2021learning}                   & 11.03 & 17.60 \\ 
        FamNet+ \cite{ranjan2021learning}                  & 11.31 & 18.84 \\
        BMNet+ \cite{shi2022represent}                   & 11.44 & 23.22 \\
        SAFECount \cite{you2023few}                & 9.80  & 32.40 \\
        Seg-Count \cite{xu2025learning}           & 6.97  & \textbf{13.03} \\
        \midrule
        OCCAM-M                  & \textbf{5.52}  & 13.43 \\
        \bottomrule
        \end{tabular}
    \label{tab:5}
\end{minipage}
\hfill
\begin{minipage}[t]{0.58\textwidth}
    \centering
    \small
    \captionof{table}{Methods on CARPK Test Set (best results in \textbf{bold}). OCCAM and Omnicount \cite{mondal2025omnicount} are Multi-Class. RCC \cite{hobley2022learning} is fine-tuned on CARPK. Methods above the double line are Training-Free, but utilize exemplars, while below are Prior-Free, but rely on training.}
        \begin{tabular}{lrrcc}
        \toprule
        Model          & MAE $\downarrow$       & RMSE    $\downarrow$    & NAE $\downarrow$  & SRE $\downarrow$  \\
        \midrule
        TFOC\cite{shi2024training}          & 10.97      & 14.24       & 0.48  & 3.70  \\
        OmniCount\cite{mondal2025omnicount}     & 9.92       & 12.15       & 0.23  & 2.11  \\
        TFCounter\cite{ting2024tfcounter}     & 9.71       & 12.44       & -     & -     \\
        ValidCounte\cite{huang2025training}  & 9.36       & 12.86       & \textbf{0.17}  & \textbf{1.41}  \\
        A-Simple-But\cite{lin2025simple}  & \textbf{4.39}       & \textbf{5.70}        & -     & -     \\
        \midrule
        \midrule
        GCA-SUN\cite{wu2024gca}        & 10.96      & 13.95       & -     & -     \\
        SAVE\cite{zgaren2025save}           & 9.56       & 19.04       & -     & -     \\
        RCC\cite{hobley2022learning}           & \textbf{9.21}      & \textbf{11.33}       & -     & -     \\
        \midrule
        OCCAM-M       & 14.83      & 21.96       & 0.21  & 2.20 \\
        \textbf{OCCAM-S}       & \textbf{10.06}      & \textbf{13.81}       & \textbf{0.13}  & \textbf{1.41} \\	
        \bottomrule
        \end{tabular}
    \label{tab:6}
\end{minipage}
\end{table*}



\begin{table*}[t]
\centering
\caption{OCCAM-S (above the double line) and OCCAM-M (below) on our proposed Synthetic Multi-class Test Set.}
    \begin{tabular}{ccccrccc}
        \toprule
        Version        & MAE $\downarrow$  & RMSE $\downarrow$  & NAE $\downarrow$  & SRE $\downarrow$ & Precision $\uparrow$ & Recall $\uparrow$ & F-1 $\uparrow$ \\
        \midrule
        1                 & 29.20 & 54.51 & 0.82 & 9.03  & 0.75      & 0.72   & 0.68 \\
        2                 & 26.56 & 53.89 & 0.87 & 10.16 & 0.78      & 0.76   & 0.72 \\
        3                 & 31.89 & 62.91 & 0.97 & 11.33 & 0.74      & 0.72   & 0.67 \\
        \midrule
        Mean              & 29.22 & 57.11 & 0.89 & 10.18 & 0.75      & 0.73   & 0.69 \\
        \midrule
        \midrule
        1                 & 21.99 & 37.45 & 0.42 & 4.17  & 0.85      & 0.59   & 0.66 \\
        2                 & 19.81 & 34.33 & 0.38 & 3.60  & 0.88      & 0.62   & 0.69 \\
        3                 & 22.28 & 39.16 & 0.44 & 4.18  & 0.85      & 0.59   & 0.66 \\
        \midrule
        Mean              & 21.36 & 36.98 & 0.41 & 3.99  & 0.86      & 0.60   & 0.67 \\
        \bottomrule
    \end{tabular}
\label{tab:7}
\end{table*}


\noindent{\bf Qualitative Results.}
We present indicative qualitative results in \cref{fig:qualres}. Tight bounding boxes are drawn on counted objects, with different colors denoting different clusters. In the top left image, our approach manages to correctly count the strawberries in a relatively complex scene. In the middle image of the top row, the depicted classes are separated properly and in the last image our model not only counts what is on the plate but also counts the green circles painted on the plate and the 4 marks of the wooden surface. Given that the images in the top row are from the FSC-147 Test Set, we observe that the single-class-per-image assumption does not hold. In the first image of the second row, the discriminative ability of our method is on display as the strawberries are grouped in two clusters based on where their cut side faces, and in the second image our approach even counts the watermarks. The third image of the row is a quite simple image from our proposed Synthetic Multi-Class Test Set consisting of 3 categories. 

Finally, on the bottom row we present some failure cases. Splitting a cluster and merging two or more clusters are typical issues which have a great effect on the final object count. In the first image of the third row, a coherent cluster, the stamps, is split, but there is some insight. Horizontal stamps are grouped together (light blue boxes), and vertical stamps are within fuchsia boxes. This issue is much more complex, since in some cases this distinction should be made. In the middle image of the third row, there is a cluster of sunglasses and a cluster of individual lenses. However, the exact objects of interest cannot be specified in our approach because it is prior-free. Thus, meaningful but not so useful clusters are sometimes formed. In the last image, it is not clear whether the objects of interest are apples or bricks on the wall and our method cannot distinguish the apples due to the perspective of the scene.

\noindent{\bf Ablation Study.}
The OCCAM-S configuration and the FSC-147 Test Set are used for the ablation study of the main components of our approach (\cref{tab:8}). Each component is important, as removing any of them degrades performance, according to the reported scores. Mask Processing is the most impactful component, followed by Clustering, which is expected for a single-class dataset with densely populated images, and given their respective stage in the pipeline. Scaling improves performance, specifically in images with tiny objects.


\begin{table*}[t]
\centering
\caption{Ablation Study of the main components of OCCAM-S configuration on FSC-147 Test Set (best results in \textbf{bold}). Columns \textbf{M}, \textbf{C}, \textbf{S} denote
\textbf{Mask Processing}, \textbf{Clustering}, and \textbf{Scaling Paradigm}, respectively.}
    \begin{tabular}{ccccccrccc}
        \toprule
        M & C & S & MAE $\downarrow$  & RMSE $\downarrow$  & NAE $\downarrow$  & SRE $\downarrow$ & Precision $\uparrow$ & Recall $\uparrow$ & F-1 $\uparrow$ \\
        \midrule
        \xmark                & \xmark          & \xmark                & 78.83 & 153.86 & 2.35 & 14.24 & 0.36      & 0.92   & 0.49 \\
        \xmark                & \xmark          & \cmark                & 78.74 & 153.74 & 2.35 & 14.23 & 0.36      & 0.92   & 0.49 \\
        \xmark                & \cmark          & \xmark                & 46.38 & 141.88 & 0.99 & 8.02  & 0.51      & 0.81   & 0.59 \\
        \xmark                & \cmark          & \cmark                & 46.19 & 141.69 & 0.99 & 8.01  & 0.51      & 0.81   & 0.59 \\
        \cmark                & \xmark          & \xmark                & 28.27 & 134.99 & 0.68 & 5.58  & 0.66      & 0.90   & 0.73 \\
        \cmark                & \xmark          & \cmark                & 25.38 & 111.10 & 0.68 & 5.29  & 0.66      & 0.90   & 0.74 \\
        \cmark                & \cmark          & \xmark                & 19.83 & 134.76 & \textbf{0.22} & 3.80  & \textbf{0.89}      & 0.81   & \textbf{0.83} \\
        \cmark                & \cmark          & \cmark                & \textbf{16.92} & \textbf{110.83} & \textbf{0.22} & \textbf{3.38}  & \textbf{0.89}      & \textbf{0.82}   & \textbf{0.83}\\
        \bottomrule
    \end{tabular}
\label{tab:8}
\end{table*}


%% file: sections/5_conclusions.tex
\section{Conclusions}
\label{chapter:conclusions}

In this paper, we introduced OCCAM, a novel approach for Class-Agnostic object Counting (CAC). It operates out-of-the-box and addresses key limitations of existing CAC methods by eliminating the need for extensive training and supplementary information, while supporting the simultaneous counting of multiple object classes. This makes OCCAM uniquely positioned, due to its minimal assumptions. By leveraging the Segment Anything Model 2 (SAM2) alongside our adapted version of the FINCH algorithm, our approach achieves competitive performance on standard benchmark datasets, such as FSC-147 and CARPK. Additionally, we proposed a synthetic multi-class dataset and emphasized the importance of adopting the F1 score as a more reliable evaluation metric in the field. We hope that our method serves as a step towards fully automated, general-purpose object counting systems applicable to real-world scenarios.
